\documentclass[12pt]{article}
\usepackage[english]{babel}
\usepackage[utf8x]{inputenc}
\usepackage{amsmath}
\usepackage{graphicx}
\usepackage[colorinlistoftodos]{todonotes}

\begin{document}

\begin{titlepage}

\newcommand{\HRule}{\rule{\linewidth}{0.5mm}} 

\center 
 

\textsc{\LARGE International Institute of Earthquake Engineering and Seismology}\\[1.5cm] 
\textsc{\Large Post-Earthquake Managements}\\[0.5cm] 
\textsc{\large Rescue Operation}\\[0.5cm] 


\HRule \\[0.4cm]
{ \huge \bfseries Resource Planning For Rescue Operations}\\[0.4cm] 
\HRule \\[1.5cm]
 

\begin{minipage}{0.4\textwidth}
\begin{flushleft} \large
\emph{Author:}\\
Eng. Mona \textsc{Khaffaf} 
\end{flushleft}
\end{minipage}
~
\begin{minipage}{0.4\textwidth}
\begin{flushright} \large
\emph{Supervisor:} \\
Eng. Arshia \textsc{Khaffaf} 
\end{flushright}
\end{minipage}\\[2cm]



{\large \today}\\[1cm] 


\includegraphics{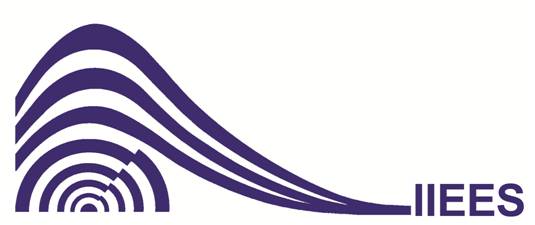} 
 

\vfill 

\end{titlepage}

\begin{abstract}
After an earthquake, disaster sites pose a multitude of health and safety concerns. A rescue operation of people trapped in the ruins after an earthquake disaster requires a series of intelligent behavior, including planning. For a successful rescue operation, given a limited number of available actions and regulations, the role of planning in rescue operations
is crucial. Fortunately, recent developments in automated planning by artificial intelligence community can help different organization in this crucial task. Due to the number of rules and regulations, we believe that a rule based system for planning can be helpful for this specific planning problem. In this research work, we use logic rules to represent rescue and related regular regulations, together with a logic based planner to solve this complicated problem. Although this research is still in the prototyping and modeling stage, it clearly shows that rule based languages can be a good infrastructure for this computational task. The results of this research can be used by different organizations, such as Iranian Red Crescent Society and International Institute of Seismology and Earthquake Engineering (IISEE). 

\end{abstract}
\pagebreak

\section{Introduction}

\begin{figure}
  \centering
  \includegraphics[scale=0.8]{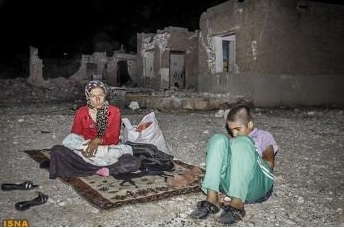}
  \caption{Earthquake survivors sit near damaged houses in the earthquake stricken town of Bushehr in Iran, April 9, 2013 ... April 11, 2013, 11:45 am EDT \cite{Reuters2013}.}
  \label{fig:damage_02}
\end{figure}

One of the greatest concerns of human being is earthquake. After an earthquake for rescue operations of people trapped in the ruins and reduction of damages, a wide series of intelligent operations in a very short time is needed. Rescue operations of people trapped in an unsafe position often associated with firefighting services which should be supported by search and rescue dogs. Moreover, some objects and tools are used might included hydraulic cutting tools, helicopters, and so on. In addition, there are several transportation issues such as moving of patients, disabled, elderly, children, prisoners, and so on, to a safe places. Moreover, to help people left in the wrecked area, a huge amount of medical helps and foods should be entered to the site (Figure~\ref{fig:damage_02}). Therefore, one of the most important factors can directly affect losses and injuries is the status of roads and transporting networks. The rescue management and operators need to properly manage the entries of cities, railways, airports, and so on. 

\begin{figure}
  \centering
  \includegraphics[scale=0.6]{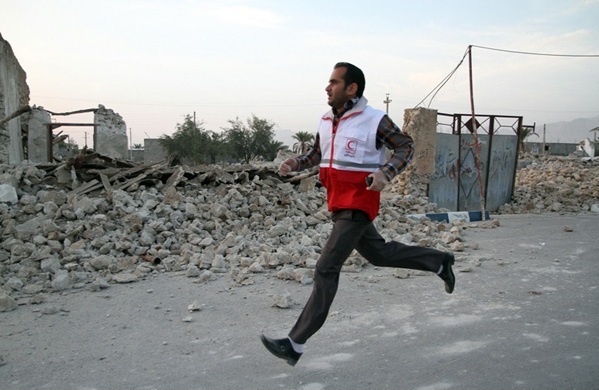}
  \caption{Rescue operation, Iran, 2013\cite{Fararu2013}.}
  \label{fig:damage_01}
\end{figure}

The rescue operation also requires building helicopter pads due to the requirements of fast transportation of injuries, management of emergency traffics. Traffic management also includes helping cars jammed in roads, bridges, and crosses. Transportation management also includes finding and creating emergency ways. Clearly, management of transportation is not the only concern in post-earthquake management. For instance, other networks such as electricity power, gas, water, and waste water can cause secondary damages and lack of these vital networks can affect life of people. However, the management of those network is not in the scope of current report.

Apparently,  in the management of this critical situation, the role of planning in rescue operations is so crucial. We believe that a rule based system for planning rescue operations can be very helpful. In this paper we establish the first, but crucial, step of our long term goal of providing a planning infrastructure with respect to existing rules and regulations. To achieve this goal, we need an expressive language to specify planning problem, i.e. actions and current state, and regulations inside a uniform deductive formalism. This way it will be possible to import the techniques of logical deduction to both planning and verification of satisfaction of our rules. In fact, this also will provide a structural analysis of plans together with our domain constraints. Our application needs to have some structure supporting rules similar to what computer programmers develop in Prolog, plus a system to express available actions and resources. 

Although PDDL 3 supports \textit{domain axioms} to derive additional information (and this can be very helpful in our case), existing non-deductive planners are not efficient enough when the number of domain axioms (rules) are growing.

Situation calculus was the first logical planning system that could incorporate logical rules \cite{Green:1969:ATP:1624562.1624585}. Unfortunately, situation calculus is a very low level language for explaining urban rules and regulations. Moreover, up to our knowledge, there is not any efficient execution engine for situation calculus. There are several planners that
encode planning problems into constraint satisfaction problems (CSP) \cite{Stefik:1980:PC:909378}\cite{DBLP:conf/aips/ErolHN94}\cite{DBLP:journals/fuin/BartakT10} and use logical deduction to solve the planning problems. It is also very difficult to translate urban regulations to a set of constraints. Answer set programming is another, more recent logic based technique to solve planning problems \cite{Lifschitz200239}\cite{DBLP:journals/etai/GelfondL98}\cite{Gebser:2012:GUE:2363344.2363364}\cite{Son:2006:DKA:1183278.1183279}. However, answer set programming also does not show a reasonable performance when the number of rules and variables increases.

Fortunately, recent developments in logical planning \cite{DBLP:conf/lpnmr/Basseda15,DBLP:conf/padl/BassedaK15,DBLP:conf/rr/BassedaK15,DBLP:conf/rr/BassedaKB14} brought hope to our camp to find a practical tool to help us in the rescue situation. However, we have not practically applied this new technique to solve our planning problem in rescue domain. The recent developments are based on a general theory, called Transaction Logic \cite{Bonner95transactionlogic,trans-chapter-98,Bonner94anoverview,trans-iclp93,concurrent-tr-96}. There is also an interpereter that can execute Transaction Logic programs \cite{Fodor:2010:TTL:1836089.1836115}. The provided technique supports rules to derive implicit information from explicitly provided facts. In our application, there are a lot factors that are dependent of existing facts, e.g. a road may be blocked if the pipes are broken and the drainage is also not working. 

The rest of this paper is structured as follows: in section 2 we discuss about planning principles in post-earthquake rescue domains, section 3 is explaining our early ideas of using Transaction Logic to solve our planning problem and the last section is summarizing our report. 

\section{Post-Earthquake Rescue Planning}

Unlike most of common industrial planning problems, physical location must be specified for almost all of objects and resources in rescue planning for post-earthquake. Such requirement makes this problem unique as most of the actions are locally defined. In order to make our problem simpler, we create an abstract graph from the detailed map of the site. The abstractness of this modeling depends on the required solution and available information.

\begin{figure}
  \centering
  \includegraphics[scale=.25]{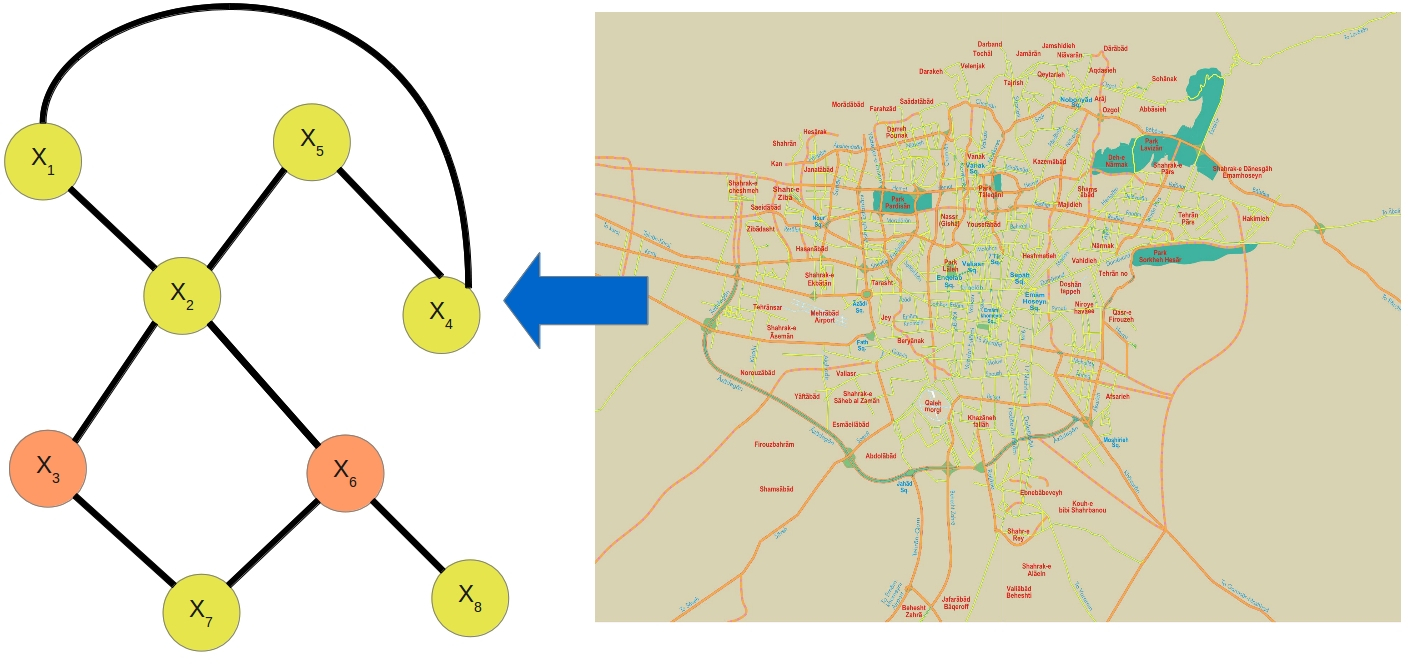}
  \caption{Creating an abstract graph from the site.}
  \label{fig:city_network}
\end{figure}

\begin{figure}
  \centering
  \includegraphics[scale=.25]{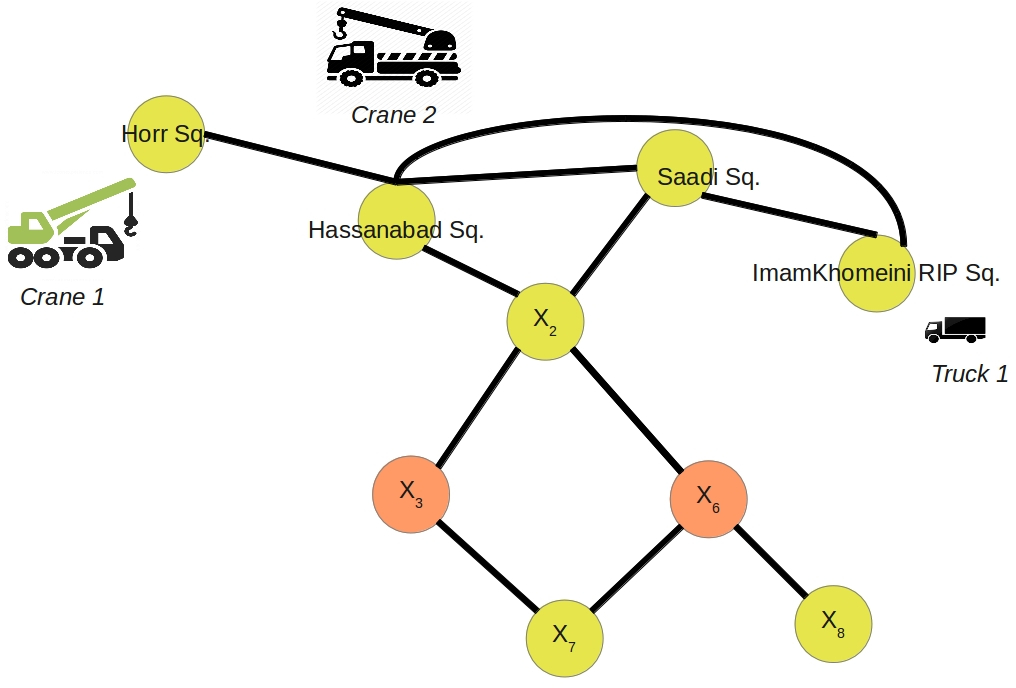}
  \caption{A sample network and resources representing the central regions of Tehran.}
  \label{fig:network_resource}
\end{figure}

In fact, our original database includes a set of tables, where each row is associated with a two dimensional coordination. Therefore, every objects and actors will be associated to the nodes representing their location areas. For example, Hassanabad Sq. of Tehran will be represented by node $ n_1 $ and all of the points considered in that region will be mapped to $ n_1 $. Then, access roads and routes can be mapped to edges. For instance, let $ n_2 $ represent Horr Sq. in Tehran. Then, there must be an edge between $ n_1 $ and $ n_2 $. Similarly, all the available resources, e.g. trucks and cranes, are bound to their locations. Figure~\ref{fig:net_rep}  shows how we represent the example graph in Figure~\ref{fig:network_resource} in terms of a set of facts. Naturally, all of this information can be represented as a set of facts. This graph not only makes the representation of problem and actions simpler, but also can reduce the complexity of the required computation. 

\begin{figure}

\begin{verbatim}
node('Horr Sq.').
node('Hassanabad Sq.').
node('Imam Khomeini RIP Sq.').
node('Saadi Sq.').
...
link('Horr Sq.','Hassanabad Sq.').
link('Saadi Sq.','Hassanabad Sq.').
link('Imam Khomeini RIP Sq.','Hassanabad Sq.').
link('Saadi Sq.','Imam Khomeini RIP Sq.').
...
crane(crane_1,big_crane).
crane(crane_2,small_crane).
truck(truck_1,mid_truck).

\end{verbatim}
\caption{Representation of network as a set of facts.}
\label{fig:net_rep}
\end{figure}

\begin{figure}
  \centering
  \includegraphics[scale=.25]{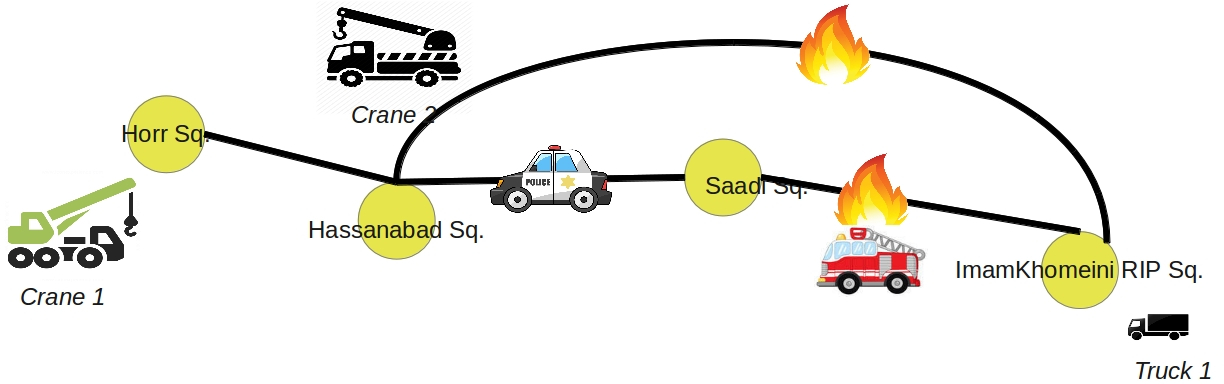}
  \caption{A sample network and hypothetical events at the central regions of Tehran.}
  \label{fig:network_events}
\end{figure}

Similarly, post-earthquake events, e.g. fire and gas leakage, also can be bound to their locations. For instance, Figure~\ref{fig:net_rep_event}  shows how we represent the post-earthquake events shown in Figure~\ref{fig:network_events} in terms of a set of facts.Since these sets of events will be updated during the rescue operations, we must be able to update these facts from observatory reports.

\begin{figure}

\begin{verbatim}
...
police_block('Saadi Sq.','Hassanabad Sq.').
fire('Imam Khomeini RIP Sq.','Hassanabad Sq.').
fire('Saadi Sq.','Imam Khomeini RIP Sq.').
fireman_operation('Saadi Sq.','Imam Khomeini RIP Sq.').
...
\end{verbatim}
\caption{Representation of network as a set of facts.}
\label{fig:net_rep_event}
\end{figure}

Finally, we define actions for each agent in terms of standard actions. Standard actions defined in IISEE can be mapped to the \emph{STRIPS} actions defined in~\cite{DBLP:conf/rr/BassedaKB14}. Each action is composed of three arguments: \emph{agent}, \emph{preconditions}, \emph{effects}. At this stage of our project, we just consider moving agents, like cranes and trucks. Therefore, the effects of our actions are simple relocation effects. The preconditions are usually more complicated. For example, we can add to the precondition of \emph{move} action of a small truck that it cannot pass a link if there exist a fire event on that link. 

The last step is translating rules and regulations. These rules are playing an important in the management of post-earthquake rescue operations. There are a lot of rules and regulations recommended (or mandated) by different organizations. Those rules are used to derive information that is used by preconditions of actions. In the extended version of our project, these rules can also be used to define new goals, checking inconsistency, hazardous situations, and etc. As an example, we define the property of $ safe $ for areas such that an area is safe if there is a link for that area without fire reported. Figure~\ref{fig:net_rep_rule} shows how we define $ safe\_area $ property in our system using $ scape\_path $ relationship. This property can be used in the preconditions of actions. For instance, we can limit $ move $ actions for small trucks such that they cannot enter an un-safe area. 

\begin{figure}
  \centering
  \includegraphics[scale=.25]{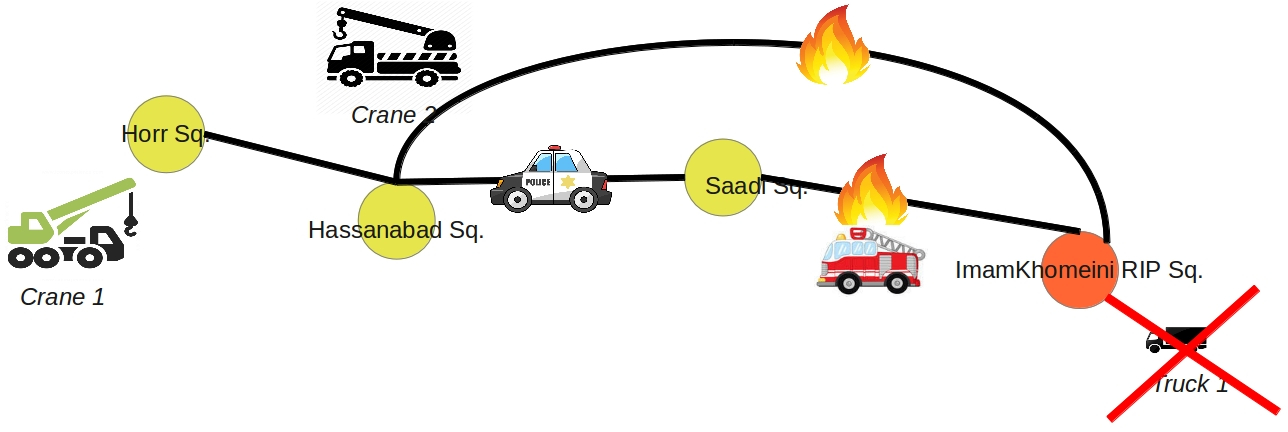}
  \caption{A sample network and events representing showing a hazardous area at the central regions of Tehran.}
  \label{fig:network_rules}
\end{figure}

\begin{figure}

\begin{verbatim}
...
scape_path(X,Y) :- link(X,Y),
                 not fire(X,Y).
scape_path(X,Y) :- link(Y,X),
                 not fire(Y,X).
safe_area(X) :- scape_path(X,_).
...
\end{verbatim}
\caption{Representation of network as a set of facts.}
\label{fig:net_rep_rule}
\end{figure}
\section{Deployment}

As mentioned in the first section, this is an ongoing project. We are planning to implement our models in a Prolog system (SICStus \cite{DBLP:journals/tplp/CarlssonM12}, SWI-Prolog \cite{DBLP:journals/tplp/WielemakerSTL12}, XSB \cite{DBLP:conf/procomet/Warren98}, or YAP-Prolog \cite{DBLP:journals/tplp/CostaRD12}). Since the existing Transaction Logic Planners \cite{DBLP:conf/lpnmr/Basseda15,DBLP:conf/padl/BassedaK15,DBLP:conf/rr/BassedaK15,DBLP:conf/rr/BassedaKB14} and their Transaction Logic interpreters  \cite{Fodor:2010:TTL:1836089.1836115} are implmented in XSB, we may use XSB for our deployment phase. The main stage of our project is translating two dimensional maps to graphs and representing them in terms of sets of facts. This phase is already finished.

\section{Summary}
We have proposed an approach for using rule-based systems for post-earthquake rescue planning and operations. Other event processing techniques \cite{DBLP:journals/semweb/AnicicRFS12,DBLP:journals/aai/AnicicRFS12,DBLP:conf/ruleml/AnicicRFS11a} may also be used for this application. This can be a feasibility study for future projects. We hope that attending RuleML 2016 can help to direct this project.

\section*{Acknowledgment}

The authors would like to thank IISEE, Iranian Red Crescent Society, and Tehran Municipality for their kind cooperation.  

\bibliographystyle{splncs}
\bibliography{RescueAgentPlanning}

\end{document}